\documentclass[11pt,a4paper]{article}

\usepackage{amsmath,amsfonts,bm}

\def\1{\bm{1}}
\newcommand{\train}{\mathcal{D}}

\def\ervc{{\textnormal{c}}}

\def\ervg{{\textnormal{g}}}

\def\ervy{{\textnormal{y}}}
\def\ervz{{\textnormal{z}}}

\def\ermC{{\textnormal{C}}}

\def\ermG{{\textnormal{G}}}

\def\vx{{\bm{x}}}

\DeclareMathAlphabet{\mathsfit}{\encodingdefault}{\sfdefault}{m}{sl}
\SetMathAlphabet{\mathsfit}{bold}{\encodingdefault}{\sfdefault}{bx}{n}

\newcommand{\R}{\mathbb{R}}

\usepackage[]{acl}
\usepackage{times}
\usepackage{xspace}
\usepackage{amsmath}
\usepackage{latexsym}
\usepackage{threeparttable}
\usepackage{multirow}
\usepackage{pbox}
\usepackage{pgfplots}
\usepackage{paralist}
\usepackage{ragged2e,array,booktabs}
\usepackage{graphicx}
\usepackage{subcaption}
\usepackage{cleveref}

\usetikzlibrary{positioning}
\usetikzlibrary{calc}
\usetikzlibrary{fit}
\usetikzlibrary{decorations.text}
\usetikzlibrary{shapes.multipart}

\usepackage{microtype}

\usepackage{xspace}
\usepackage{threeparttable}
\usepackage{multirow}
\usepackage{graphicx}
\usepackage{tabularx}
\usepackage{mathtools}
\usepackage{amsmath}
\usepackage{booktabs}
\usepackage{multicol}
\usepackage{multirow}
\usepackage{dsfont}
\usepackage{amsmath}

\newcommand{\secref}[2][]{Section#1~\ref{#2}\xspace}

\newcommand{\xd}[1]{#1}
\newcommand{\cameraReady}[1]{#1}

\newcommand{\aae}{AAE\xspace}
\newcommand{\sae}{SAE\xspace}
\newcommand{\wrt}{\emph{w.r.t.}\@\xspace}

\newcommand{\class}[1]{\textsc{#1}\xspace}
\newcommand{\happy}{\class{happy}}
\newcommand{\sad}{\class{sad}}
\newcommand{\dataset}[1]{\textbf{#1}\xspace}

\newcommand{\moji}{\dataset{Moji}}
\newcommand{\bios}{\dataset{Bios}}

\newcommand{\bert}{BERT\xspace}
\newcommand{\deepmoji}{DeepMoji\xspace}

\newcommand{\method}[1]{\textsf{#1}\xspace}
\newcommand{\vanilla}{\method{Vanilla}}
\newcommand{\adv}{\method{ADV}}
\newcommand{\dadv}{\method{DADV}}
\newcommand{\inlp}{\method{INLP}}

\newcommand{\ds}{\method{DS}}

\newcommand{\btrw}{\method{RW}}
\newcommand{\rw}{\method{RW}}

\newcommand{\eo}{\method{EO}}
\newcommand{\difference}{\method{EO$_\mathrm{{CLA}}$}}
\newcommand{\mean}{\method{EO$_\mathrm{{GLB}}$}}
\newcommand{\fairbatch}{\method{FairBatch}}

\newcommand{\accuracy}{Accuracy\xspace}

\newcommand{\gap}{GAP\xspace}

\newcommand{\dto}{DTO\xspace}
\newcommand{\fairness}{Fairness\xspace}

\newcommand{\pearson}{Pearson\xspace}
\newcommand{\pearsongap}{Pearson-GAP\xspace}
\newcommand{\mae}{MAE\xspace}
\newcommand{\maegap}{MAE-GAP\xspace}
\newcommand{\rmse}{RMSE\xspace}
\newcommand{\rmsegap}{RMSE-GAP\xspace}

\title{Systematic Evaluation of Predictive Fairness}

\author{Xudong Han$^{\spadesuit}$\quad
        Aili Shen$^\heartsuit$\thanks{~~This work was done when Aili Shen was at The University of Melbourne.}\quad
    	Trevor Cohn$^\spadesuit$\quad
    	Timothy Baldwin$^{\spadesuit\diamondsuit}$\quad
    	Lea Frermann$^\spadesuit$ \\
        $\spadesuit$ School of Computing and Information Systems, The University of Melbourne \\
        $\heartsuit$ Amazon Alexa AI, Australia \\
        $\diamondsuit$ Department of Natural Language Processing, MBZUAI\\
	    \url{xudongh1@student.unimelb.edu.au}, 
	    \url{aili.shen@amazon.com} \\
	    \url{{t.cohn,tbaldwin,lfrermann}@unimelb.edu.au} 
  }

\date{}

\begin{document}

\maketitle

\begin{abstract}
Mitigating bias in training on biased datasets is an important open problem. Several techniques have been proposed, however the typical evaluation regime is very limited, considering very narrow data conditions. For instance, the effect of target class imbalance and stereotyping is under-studied.
%
%
To address this gap, we examine the performance of various debiasing methods across multiple tasks, spanning binary classification (Twitter sentiment), multi-class classification (profession prediction), and regression (valence prediction). Through extensive experimentation, we find that data conditions have a strong influence on relative model performance, and that general conclusions cannot be drawn about method efficacy when evaluating only on standard datasets, as is current practice in fairness research. {\it Our code is available at: \url{https://github.com/HanXudong/Systematic_Evaluation_of_Predictive_Fairness}.}
\end{abstract}

\section{Introduction and Background}
\label{introduction}

Naively-trained models have been shown to encode and amplify biases in the training dataset, and exhibit performance disparities across author demographics~\citep{Hovy:15,Li:18,Wang:19}.
Various methods have been proposed to mitigate such biases, such as balanced training~\citep{Zhao:18,Han:21c}, adversarial debiasing~\citep{Elazar:18,Han:21}, and null-space projection~\citep{Shauli:20, ravfogel2022linear}.
However, experiments have largely been conducted on a handful of benchmark datasets such as  \moji sentiment analysis~\citep{Blodgett:16} and \bios biography classification~\citep{De-Arteaga:19}, under a narrow set of data conditions.

In this paper, we systematically explore the impact of data conditions on model accuracy and fairness, synthesising the following data conditions over real-world datasets: (1) target label (im)balance; (2) protected attribute (im)balance; (3) target label--protected attribute (im)balance (also known as ``stereotyping''); and (4) target label arity. Consistent with the literature on fairness in NLP, we primarily focus on classification tasks, but also include preliminary text regression experiments. In doing so, we develop a novel framework for comprehensively evaluating the performance of debiasing methods under a range of data conditions, and use it to evaluate eight widely-used debiasing methods.

Our experimental results show that there is no single best model. Debiasing methods that account for both class disparities and demographic disparities are generally more robust, but are less effective in multi-class settings. 
For the regression task, our experiments indicate that existing debiasing approaches can substantially improve fairness, and that simple linear debiasing outperforms more complex methods.



\section{Related Work}


In this section, we first describe different fairness criteria, then examine work which has evaluated the effectiveness of debiasing methods from different perspectives. 

\paragraph{Fairness Criteria}
Studies in the fairness literature have proposed several definitions of fairness capturing different types of discrimination, such as group fairness \cite{Hardt:16,Zafar:17,Cho:20,Zhao:20}, individual fairness \cite{Saeed:19,Yurochkin:20,Dwork:12}, and causality-based fairness \cite{Wu:19,Zhang:18,Zhang:18b}. In this work, we focus on group fairness, where a model is considered to be fair if it performs identically across different demographic subgroups. 

To quantify how predictions vary across different demographic subgroups, demographic parity \cite{Feldman:15,Zafar:17b,Cho:20}, equal opportunity \cite{Hardt:16,Madras:18}, and equalized odds \cite{Cho:20,Hardt:16,Madras:18} are widely-used notions. We present these in a setting where there are exactly two protected attribute labels (a ``privileged'' and ``under-privileged'' sub-population), consistent with how they are traditionally defined. \textit{Demographic parity} ensures that models achieve the same positive prediction rate for the two demographic subgroups, 
not taking the ground-truth target label into consideration. \textit{Equal opportunity} requires that models achieve the same true positive rate across the two subgroups for instances with a positive label. \textit{Equalized odds} goes one step further in requiring that  models achieve not only the same level of true positive rate but also the same level of false positive rate across the two groups. 

\cameraReady{
Aligned with key applications such as loan approvals, most fairness metrics
assume binary classification and focus on one label (e.g., loan approved.)
When turning attention to a multi-class classification scenario,
\emph{equal opportunity} is a natural choice, as it can be easily reformulated by assigning the positive class to each candidate class under a 1-vs-rest formulation.
}

\paragraph{Effectiveness of Debiasing Methods}

Beyond the standard definitions of fairness, a number of studies have examined the effectiveness of various debiasing methods in additional settings \cite{Gonen:19,Meade:21,Lamba:21,Baldini:22,Chalkidis:22}. For example, 
\citet{Meade:21} not only examine the effectiveness of various debiasing methods but also measure the impact of debiasing methods on a model's language modeling ability and downstream task performance. \citet{Kellie:20} find that existing pretrained models encode different degrees of gender correlations, despite their performance on target tasks being quite similar, motivating the need to consider different metrics when performing model selection. A similar effect is also observed by \citet{Baldini:22}. \citet{Chalkidis:22} examine the effectiveness of debiasing methods over a multi-lingual benchmark dataset consisting of four subsets of legal documents, covering five languages and various sensitive attributes. They find that methods aiming to improve worse-case performance tend to fail in more realistic settings, where both target label and protected attribute distributions vary over time. \citet{Lamba:21} perform an empirical comparison of various debiasing methods in solving real-world problems in high-stakes settings, all of which take the form of binary classification tasks. However, the effectiveness of debiasing methods under different data distributions (in terms of target class and protected attribute) has not been systematically investigated.

\section{Methods}
Here we describe the methods employed to manipulate the dataset distributions for classification tasks, and then describe how we adopt debiasing methods to a regression setting. 

\subsection{Notation Preliminaries}
Experiments are based on a dataset consisting of $n$ instances $\train=\{(\vx_{i}, \ervy_{i}, \ervz_{i})\}_{i=1}^{n}$, where $\vx_{i}$ is an input vector, $\ervy_{i}\in \{ \ervc \}_{ \ervc = 1 }^{\ermC}$ represents target class label, and $\ervz_{i} \in \{ \ervg \}_{ \ervg = 1 }^{\ermG}$ is the group label, such as gender. 
$n_{\ervc, \ervg}$ denotes the number of instances in a subset with target label $\ervc$ and protected label $\ervg$, i.e., $\train_{\ervc, \ervg}=\{(\vx_{i}, \ervy_{i}, \ervz_{i})|\ervy_{i}=\ervc, \ervz_{i}=\ervg\}_{i=1}^{n}$.
The corresponding empirical probability of combination of $\ervy$ and $\ervz$ values is $P(\ervy = \ervc, \ervz = \ervg) = \frac{n_{\ervc, \ervg}}{n}$.


\subsection{Manipulating Label Distributions}
\label{sec:manipulating_distributions}

To investigate the effectiveness of debiasing methods under different data distributions, we need the ability to create synthetic datasets $\train'$ that follow arbitrary distributions $P'(\ervy, \ervz)$.
Intuitively, given $m$ instances and the joint probability \mbox{$P'(\ervy = \ervc, \ervz = \ervg)$}, we can create each of the subsets $\train'_{\ervc, \ervg}$ by sampling 
$mP'(\ervy = \ervc, \ervz = \ervg)$ instances with replacement from $\train_{\ervc, \ervg}$.
However, each $P'$ has $\ermC\times\ermG$ parameters, rendering a systematic analysis infeasible. Instead,  we propose to control the joint distribution in an interpretable way, via a single parameter\xd{, and report results as graphs}:
Given a particular rate $0 \leq \alpha \leq 1 $, we define the arbitrary distribution $P'(\ervy, \ervz)$ as the interpolation between the empirical distribution $P(\ervy, \ervz)$ and a distribution of interest $Q(\ervy, \ervz)$:
$$P'(\ervy, \ervz) = (1-\alpha) P(\ervy, \ervz) + \alpha Q(\ervy, \ervz).$$
Next, we adopt two balanced training objectives~\citep{Han:21c} as our $Q$ distributions, and discuss their relationship to fairness.


\textbf{Conditional Balance (CB)} follows the notion of equal opportunity and emphasises the balance of demographics within each class, i.e., $Q_{\text{CB}}(\ervz=\ervg|\ervy=\ervc)=\frac{1}{\ermG}, \forall \ervg \in \{1,\dots,\ermG\}, \ervy \in \{1,\dots,\ermC\} $. The resulting interpolation is:
\begin{equation*}
    P_{\text{CB}}'(\ervy, \ervz) 
     =  P(\ervy) [(1-\alpha) P(\ervz|\ervy) + \alpha Q_{\text{CB}}(\ervz|\ervy) ]
\end{equation*}
where the overall class distribution $P(\ervy)$ does not change with the value of $\alpha$. 

\textbf{Joint Balance (JB)} goes one step further in taking both class balance and demographic balance into account, resulting in $Q_{\text{JB}}(\ervz=\ervg,\ervy=\ervc)=\frac{1}{\ermC \ermG}, \forall \ervg \in \{1,\dots,\ermG\}, \ervy \in \{1,\dots,\ermC\} $.
The interpolation
\begin{equation}
    P_{\text{JB}}'(\ervy, \ervz) =  (1-\alpha) P(\ervy, \ervz) + \alpha Q_{\text{JB}}(\ervy, \ervz) 
    \label{eq:jb_interpolation}
\end{equation}
ensures both class and demographic labels are more balanced with a larger $\alpha$.

\textbf{Inverting the Bias}
$\alpha=0$ and $\alpha=1$ result in the original distribution and a balanced distribution, respectively. 
{We extend the space of possible distributions, by also considering scenarios with $\alpha>1$, which result in ``anti-stereotypical'' distributions where majority classes and demographics are swapped to minorities.}

\xd{Although the sum of adjusted probabilities is guaranteed to be $1$, it is possible to generate negative probabilities or values that are larger than $1$ after interpolation. 
In \Cref{sec:appendix_Normalization_For_Probability_Table}, we describe the normalisation strategies to get a valid probability table.} 
In this paper, we consider $\alpha \in [0,2]$ for our dataset interpolations.
\xd{
Taking the CB interpolation as an example, given $P(\mathrm{Female}|\mathrm{Nurse})=0.9$ (\Cref{sec:appendix_bios_distribution}),  $\alpha = 0, 1, \mathrm{and}~2$ result in the adjusted $P'(\mathrm{Female}|\mathrm{Nurse}) = 0.9, 0.5, \mathrm{and}~0.1$, respectively. Consistent adjustments will be applied to other professions in the training dataset.
}

\subsection{Debiasing for Regression Tasks}
\label{sec:methods_regression}
Regression models predict a real-valued target variable, rather than discrete values as in classification. 
Many existing fairness metrics and debiasing methods assume discrete target (and protected attribute) labels, and are thus not directly applicable to regression tasks, such as the equal opportunity criteria which measures disparities across demographics within each class 
\citep{Roh:21,Shen:22}.

As a first step towards applying debiasing methods to text regression tasks, we map the continuous target variables $\ervy$ into discrete values by approximating the real-valued outputs with quantile-based proxy labels $\tilde{\ervy}$. Specifically, let $\tilde{\ervy}$ denote the proxy label, such that the dataset for regression is $\train=\{(\vx_{i}, \ervy_{i}, \ervz_{i}, \tilde{\ervy}_{i})\}_{i=1}^{n}$, where $\ervy\in\R$ is the continuous target label.
Given a particular number of quantiles $\tilde{\ermC}$, $\ervy$ is converted into equal-sized buckets based on sample quantiles, resulting in categorical proxy labels $\tilde{\ervy}\in\{\tilde{\ervc}\}_{\tilde{\ervc}=1}^{\tilde{\ermC}}$.
Two typical choices for $\tilde{\ermC}$ are 10 and 4, corresponding to deciles and quartiles, respectively.

In model training, we calculate losses based on real labels $\ervy$, and identify protected groups based on $\tilde{\ervy}$. Appendix~\ref{sec:appendix_regression_adaptation} presents further details for adopting debiasing methods to regression tasks.

\section{Experiments}
In this section we describe general settings across all experiments. In Appendix~\ref{sec:implementation_details}, we provide full experimental details and dataset statistics.

\subsection{Debiasing Methods}

Our focus in this work is to examine the effectiveness of various debiasing methods on different dataset compositions and their applicability to regression tasks. As such, we take a representative sample of debiasing methods, populating the spectrum of pre-processing, in-processing, and post-processing approaches. 

\paragraph{Vanilla:} The model is trained naively with cross-entropy loss, without taking bias mitigation into consideration (\vanilla). 

\paragraph{Pre-processing:} perform downsampling or reweighting of the dataset before model training.  
\begin{compactenum}
	\item Downsampling (\ds: \citet{Han:21c}): Bias mitigation is achieved by downsampling the dataset, by balancing it \wrt the protected attribute within each target class while preserving the original target class ratio. 
	\item Reweighting (\btrw: \citet{Han:21c}): Bias mitigation is achieved by assigning different weights to instances in the dataset, by reweighting based on the (inverse) of the joint distribution of the protected attribute and target classes. 
\end{compactenum}

\paragraph{In-processing:} perform adversarial training or directly optimise \wrt fairness criteria by either dynamically adjusting the sampling rate or penalising groups of instances.

\begin{compactenum}
	\item Adversarial training (\adv: \citet{Elazar:18,Li:18}) jointly trains a discriminator to predict the protected attribute, leading to representations agnostic to protected attributes. 
	\item Diverse adversarial training (\dadv: \citet{Han:21}) trains multiple discriminators as above, with a pairwise orthogonality constraint over discriminators to encourage  learning of different representational aspects.
	\item Fair batch selection (\fairbatch: \citet{Roh:21}) dynamically adjusts the instance resampling probability during training \wrt a given target class and protected attribute value, based on the equal opportunity criterion.
	\item Equal opportunity (\eo: \citet{Shen:22}) directly optimises for equal opportunity by penalising loss differences across protected groups via a regularisation term. We adopt two versions of optimising equal opportunity: {enforcing equal opportunity by aligning group-wise losses within each class (\difference), and enforcing equal opportunity globally by aligning class- and group wise loss with the overall model performance (\mean).}
\end{compactenum}




\paragraph{Post-processing:} manipulate the learned representations to achieve fairness.
\begin{compactenum}
\item Iterative null-space projection (\inlp: \citet{Shauli:20}) first
  learns dense representations with a cross-entropy loss, and then
  iteratively projects the representations to the null-space of discriminators for the protected attributes. 
\end{compactenum}

\subsection{Evaluation Metrics}

To evaluate model performance, we adopt \accuracy in our classification experiments, and Pearson correlation for the regression task.

To measure bias, following previous studies \cite{De-Arteaga:19,Shauli:20,Shen:22}, we adopt root mean square of true positive rate gap over all classes (\gap), which is defined as $\text{\gap}=\sqrt{\frac{1}{\ermC}\sum_{\ervy}(\mathrm{GAP}^{\mathrm{TPR}}_{\ervy})^{2}}$. Here, $\mathrm{GAP}^{\mathrm{TPR}}_{\ervy}=|\mathrm{TPR}_{\ervy,\ervz}-\mathrm{TPR}_{\ervy, \neg \ervz}|, \forall\ervy$, and $\mathrm{TPR}_{\ervy,\ervz}=\mathds{P}\{\hat{\ervy}=\ervy|\ervy,\ervz\}$, indicating the percentage of correct predictions among instances with the target class $\ervy$ and protected attribute label $\ervz$. $\mathrm{GAP}^{\mathrm{TPR}}_{\ervy}$ measures the absolute performance difference between demographic subgroups conditioned on target label $\ervy$, and a value of $0$ indicates that the model makes predictions independent of the protected attribute. To be consistent with our performance evaluation metrics (the higher the better), we define \fairness as $1-$\gap, where a value of 1 indicates there is no predictive bias.  


\subsection{Experimental Setup}

For each dataset, we vary training set distributions while keeping the test set fixed.
Document representations are first obtained from the given pretrained model without finetuning. Then document representations are fed into two feed-forward layers with a hidden size of 300, each followed by the {$\tanh$} activation function. We use {A}dam \cite{Kingma:14} to optimise the model for at most 100 epochs with early stopping, where training is stopped if no improvement is observed over the dev set for 5 epochs. 
All models are trained and evaluated on the same dataset splits, and models are selected based on their performance on the development set, as described in Section~\ref{ssec:modelselection}. All experiments are conducted with the {\it fairlib} library~\citep{Han:22}.


\subsection{Model Selection}
\label{ssec:modelselection}
Simultaneously optimising models for performance and fairness is a multi-objective problem, making model selection a non-trivial task. In this work, following \citet{Han:21c}, we perform model selection based on Distance to the Optimal point (\dto), where the optimal point represents the highest theoretical performance and fairness level any model can achieve. \dto supports the comparison of models by aggregating performance and fairness into a single figure of merit, where lower is better.

\section{Binary Classification}
\label{sec:binary_classification}
The task is to predict the binary sentiment (\happy and \sad) of a given English tweet, as determined by the (redacted) emoji used in the tweet. Each tweet is also associated with a binary protected attribute, reflecting the ethnicity of the tweet author, as captured in the register of the English: Standard American English (\sae) and African American English (\aae).

We use the widely-used Twitter emoji dataset \cite{Blodgett:16,Shauli:20,Shen:22}, denoted as \moji. 
The training dataset is balanced in terms of both sentiment and ethnicity in general, but skewed in terms of sentiment--ethnicity combinations, $P(\aae|\happy) = P(\sae|\sad)=0.8$.\footnote{The dev and test set are balanced in terms of sentiment–ethnicity combination.}
Due to the fact the the original dataset has been balanced with respect to targets and demographics, the CB interpolation is exactly the same as the JB interpolation (\Cref{sec:manipulating_distributions}). 


For ease of comparison with previous work~\citep{Subramanian:21}, we refer to the CB interpolation as varying ``stereotyping'' ($P'(\ervz|\ervy)$) with balanced target class distribution.
To explore the effects of target class distribution and stereotyping, we further experiment in various controlled settings: (1) varying class ratio ($P'(\ervy)$) without stereotyping ($P'(\ervz|\ervy) = 0.5$); (2) varying stereotyping with imbalanced target class distribution; and (3) varying class ratio with stereotyping. Finally, we summarise our findings with respect to the effectiveness and robustness of various debiasing methods over different class-stereotyping compositions.


\begin{figure}[t!]
	\centering
	\includegraphics[width=\linewidth]{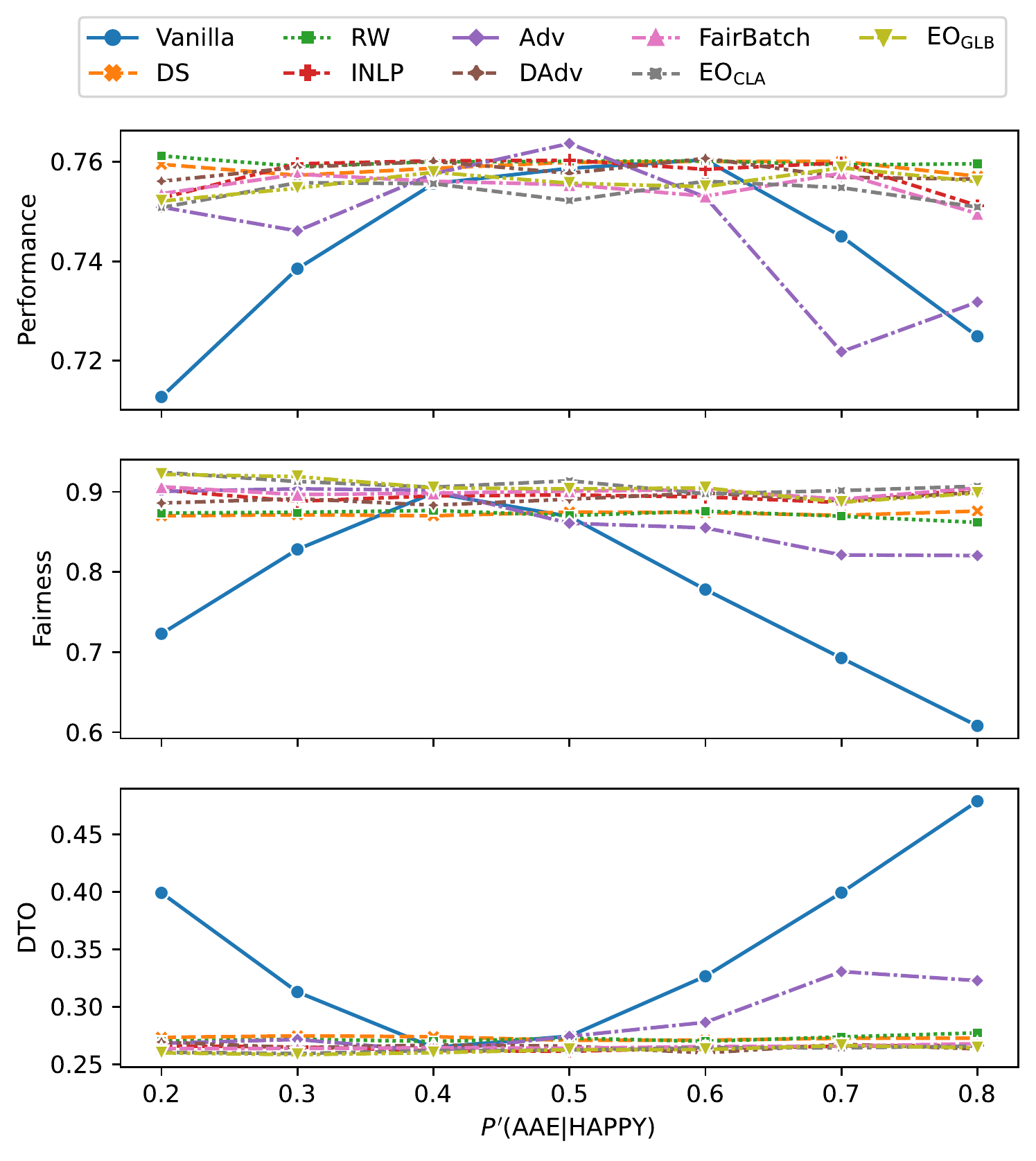}
	\caption{Results for \moji when varying $P'(\aae|\happy)$ with $P'(\happy)=P'(\sad)$. }
	\label{class_ratio_5_gender}
\end{figure}

\subsection{Varying Stereotyping with Balanced Class Distribution (CB Interpolation)}
\label{stereotyping_balanced}

Here, both sentiment and ethnicity are balanced, but skewed in terms of $P'(\aae|\happy)$ and $P'(\sae|\sad)$, ranging from 0.2 to 0.8. For example, when the ratio of \aae is 0.2, the training data composition is 10\% \happy--\aae, 40\% \happy--\sae, 40\% \sad--\aae, and 10\%~\sad--\sae. 

\Cref{class_ratio_5_gender} shows model performance in terms of \accuracy, \fairness, and \dto. All models except for \vanilla and \inlp perform similarly over varying degrees of stereotyping across metrics, indicating that most models are robust to different degrees of stereotyping using the proposed model selection approach. Turning to \vanilla, we find that \accuracy, \fairness, and \dto all vary greatly as we increase the degree of stereotyping, indicating that stereotyping affects naively-trained models in terms of both performance and fairness. 

\begin{figure}[t!]
	\centering
	\includegraphics[width=\linewidth]{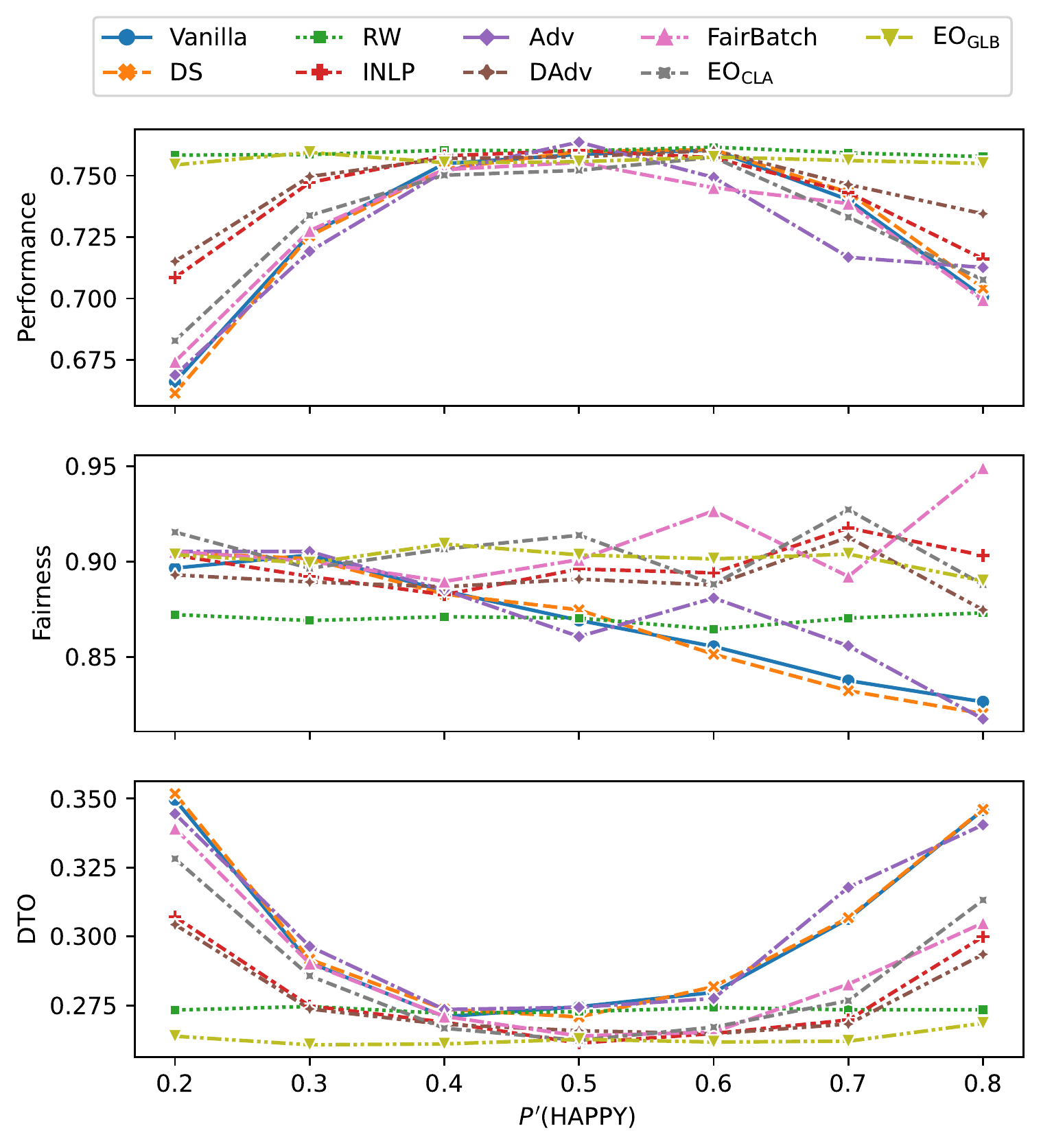}
	\caption{Results for \moji when varying $P'(\happy)$ with $P'(\aae|\happy)=P'(\sae|\sad)=0.5$. }
	\label{gender_ratio_5_class}
\end{figure}

\subsection{Varying Class Ratio with no Stereotyping}
\label{class_no_stereotyping}

In this setting, $P'(\aae|\ervy) = P'(\sae|\ervy), \forall \ervy$, and we vary $P(\ervy=\happy)$ from 0.2 to 0.8. For example, when the ratio of \happy is 0.2, the training dataset contains 10\% \happy--\aae, 10\% \happy--\sae, 40\% \sad--\aae, and 40\% \sad--\sae. 

From \Cref{gender_ratio_5_class}, we can see that most models are sensitive to the target class distribution, especially in terms of \accuracy and \dto. \btrw and \mean are exceptions, and are clearly superior methods  when the dataset is free of stereotyping, no matter the target class distribution. 
The \fairness achieved 
\xd{by all models}
in this setting does not vary greatly (ranging from approximately $0.82$ to $0.90$), indicating that target class distributions with no stereotyping have limited effect in biasing naively-trained models. 

\begin{figure}[t!]
	\centering
	\includegraphics[width=\linewidth]{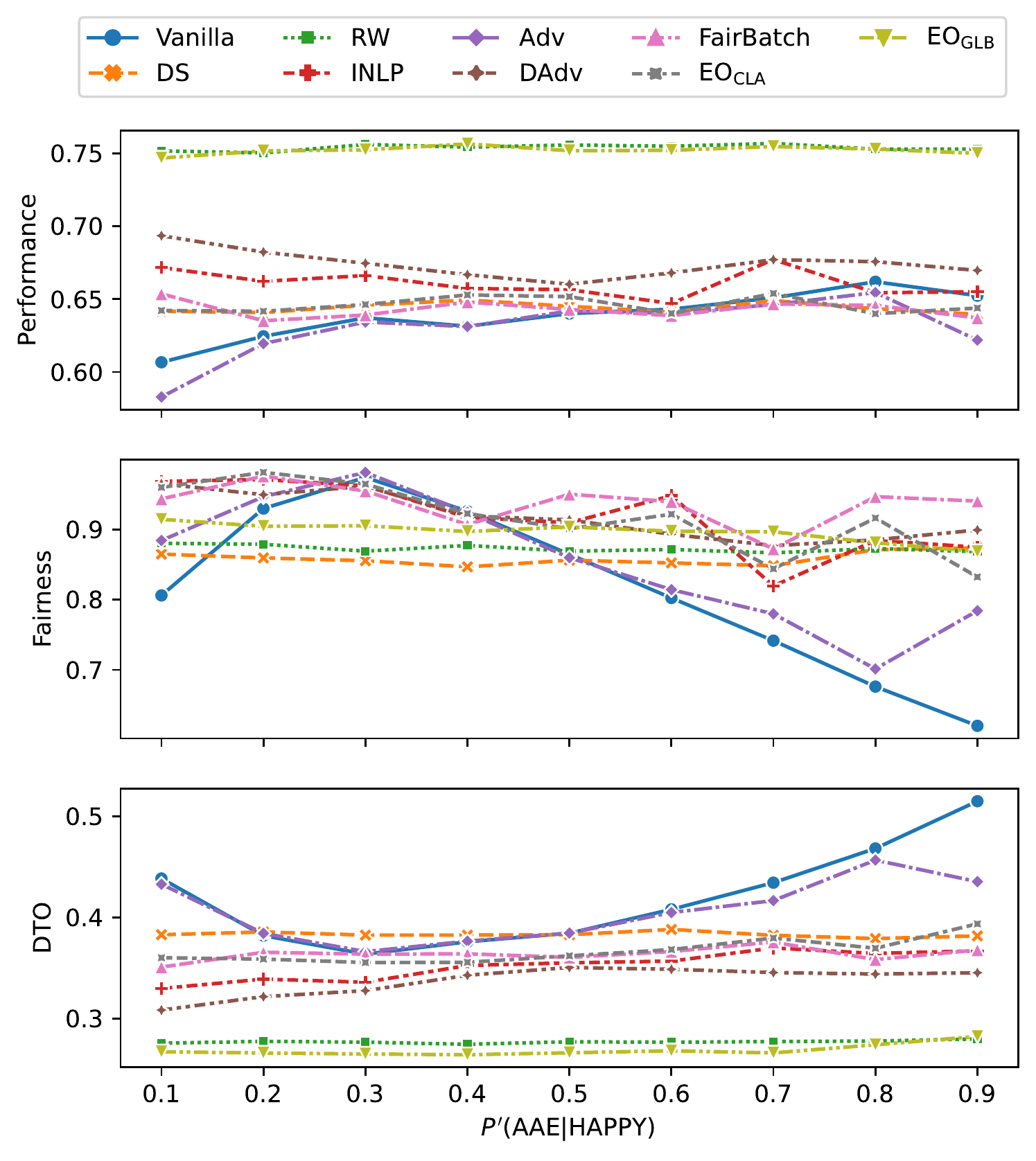}
	\caption{Results of varying $P'(\aae|\happy)$ with $P'(\happy)=0.9$. }
	\label{class_ratio_9_gender}
\end{figure}

\subsection{Varying Stereotyping with Imbalanced Class Distributions}
\label{stereotyping_imbalanced}

In this setting, the target class distribution is imbalanced, in that  $P'(\happy) = 0.9$ in the training dataset. $P'(\aae|\happy)$ and $P'(\sae|\sad)$ varies from 0.1 to 0.9. For example, when the ratio of \aae is 0.2, the training dataset contains 18\% \happy--\aae, 72\% \happy--\sae, 8\% \sad--\aae, and 2\% \sad--\sae, respectively.

From \Cref{class_ratio_9_gender}, we can see that \btrw and \mean consistently achieve the best performance in terms of \accuracy and \dto. \fairness for \ds, \rw, and \mean is robust to varying degrees of \aae stereotyping, while the remaining methods are sensitive to stereotyping.  

\begin{figure}[t!]
	\centering
	\includegraphics[width=\linewidth]{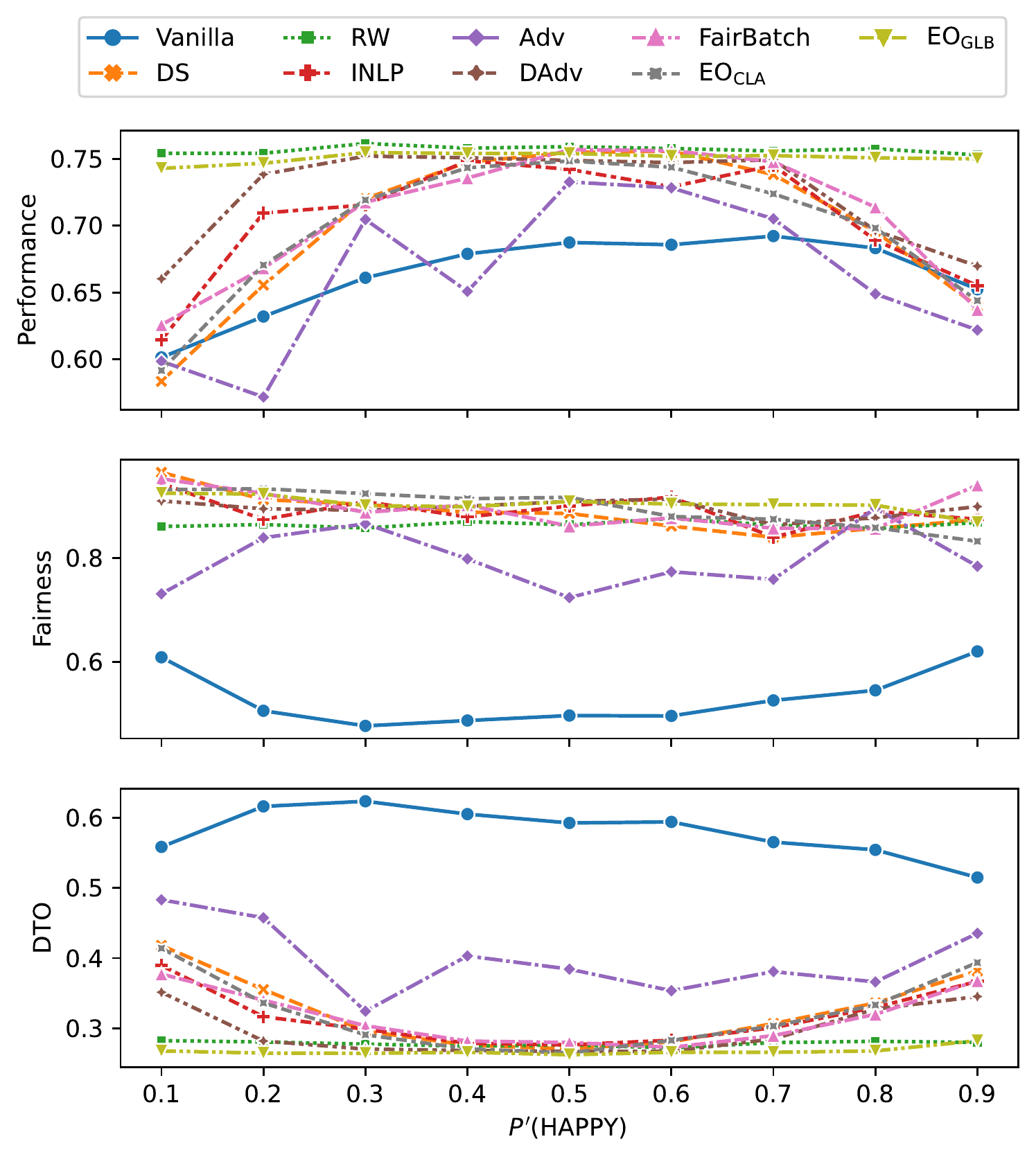}
	\caption{Results for \moji when varying $P'(\happy)$ with $P'(\aae|\happy)=P'(\sae|\sad)=0.9$. }
	\label{gender_ratio_9_class}
\end{figure}

\subsection{Varying Class Ratio with Stereotyping}
\label{class_stereotyping}

In this setting, the ethnicity distribution is imbalanced, in that $P'(\aae|\happy) = P'(\sae|\sad) = 90\%$. $P'(\happy)$ varies from 0.1 to 0.9. For example, when the ratio of \happy is 0.2, the training dataset consists of  18\% \happy--\aae, 2\% \happy--\sae, 8\% \sad--\aae, and 72\% \sad--\sae, respectively.

From \Cref{gender_ratio_9_class}, we can see that both \btrw and \mean consistently achieve the best performance in terms of \accuracy and \dto, while the remaining methods are quite sensitive to the target class distribution in terms of \accuracy and \dto, and all models except for \vanilla and \inlp achieve relatively consistent \fairness. 

\subsection{Summary}

In this section, we have performed various experiments on the Twitter sentiment analysis task with varying dataset composition. 
Looking at results from \secref[s]{stereotyping_balanced} and \ref{stereotyping_imbalanced}, we can see that all models except for \vanilla and \inlp are quite consistent with respect to \accuracy, \fairness, and \dto, with \btrw and \mean consistently achieving competitive performance in terms of \accuracy, \fairness, and \dto. Comparing results from \secref[s]{class_no_stereotyping} and \ref{class_stereotyping}, the performance of all models except for \btrw and \mean vary with respect to the target class distribution in terms of \accuracy and \dto, while all models perform consistently in terms of \fairness.

\section{Multi-class Classification}
\label{sec:profession_classification}

We next turn to our second dataset, which is a {\it multi-class classification} task with {\it natural imbalance} in both target labels and protected groups.


The dataset consists of online biographies, labeled with one of 28 occupations (target labels) and binary author gender (protected label), and the task is to predict the occupation from the biography text~(\bios, \citet{De-Arteaga:19}).

\subsection{Results}
\begin{figure}[t!]
	\centering
	\includegraphics[width=\linewidth]{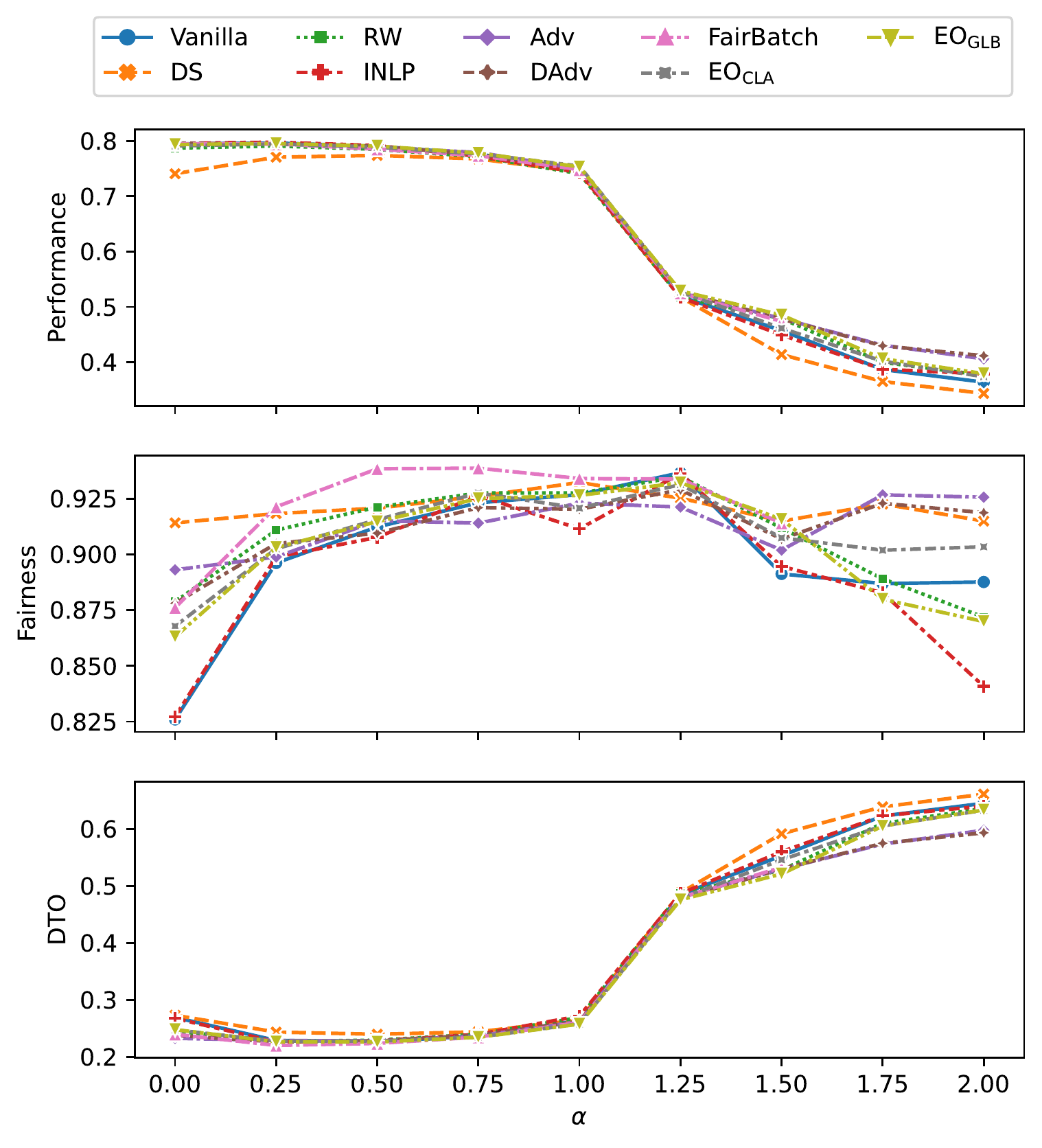}
	\caption{Results for \bios when varying the interpolation ratio under JB. Target classes and demographics are jointly balanced at $\alpha=1$.}
	\label{bios_dist_joint}
\end{figure}

\begin{figure}[t!]
	\centering
	\includegraphics[width=\linewidth]{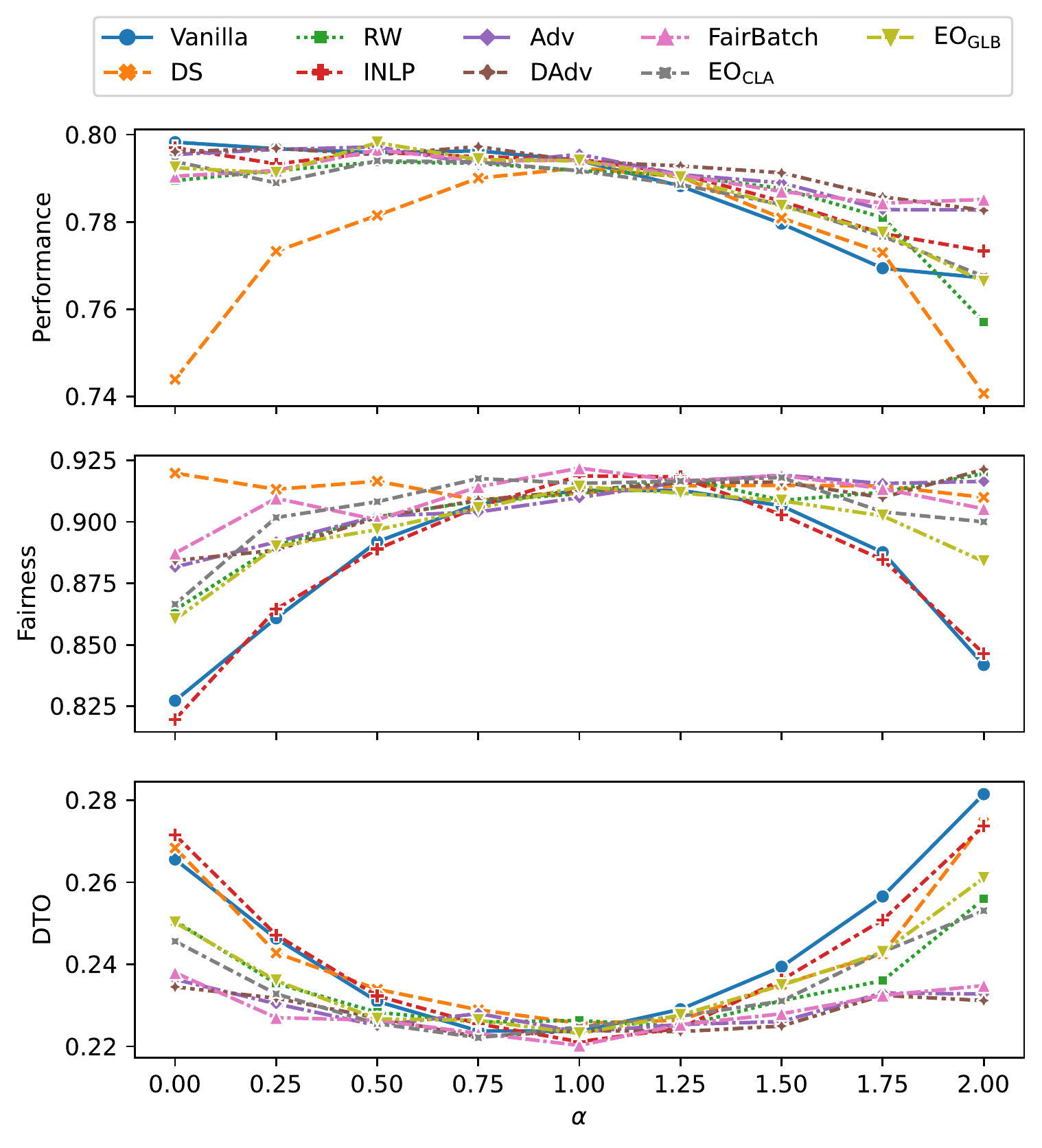}
	\caption{Results for \bios when varying the interpolation ratio under CB. Stereotyping ratios are balanced for the the $\alpha=0$ setting.}
	\label{bios_dist_stereotyping}
\end{figure}

Figures~\ref{bios_dist_joint} and~\ref{bios_dist_stereotyping} present results for JB and CB interpolation over \bios.
\xd{As introduced in \Cref{sec:manipulating_distributions}, 
JB jointly adjusts the extent of stereotyping and target class imbalance, and CB focuses on the stereotyping.}

\paragraph{JB Interpolation:}
As the value of $\alpha$ increases from 0 to 1, the training distribution becomes more balanced for both class and protected attributes, resulting in fairness improvements.
As the performance is measured as the overall accuracy, which is essentially a micro-average and oblivious to class balance, the overall performance does not improve with a more balanced class distribution.

With the $\alpha$ value further increasing from 1 to 2, both class and protected attribute distributions are biased in the opposite direction, i.e., majority groups become minority groups.
As a result,  the fairness for \vanilla decreases substantially.
Recall that the {\it test} dataset distribution is unchanged throughout the experiments (and has an identical distribution to the $\alpha=0$ setting), leading to large drops in performance of models {\it  trained} on anti-biased class distributions.

Consistent with Sections \ref{stereotyping_imbalanced}~and~\ref{class_stereotyping}, \mean outperforms other debiasing methods when the class and protected attributes are both imbalanced, as it explicitly mitigates both biases simultaneously.

We notice that \fairbatch relies on a large number of instances per class/group combination for effective resampling, and as a result is highly vulnerable to input data bias, which can be seen in the fact that there are no results for \fairbatch in imbalanced settings ($\alpha=1.75$ and $2$).\footnote{See \Cref{sec:resampling_and_reweighting_methods} for further discussion.}


\paragraph{CB Interpolation:}
When focusing on stereotyping, different methods achieve similar performance except for \ds, due to the simple sampling strategy substantially reducing the training dataset size.

In terms of \fairness, debasing approaches except for \inlp are robust to different stereotyping levels.
\mean achieves worse performance than \difference because it additionally considers class imbalance. 
As \adv and \dadv mitigate biases without taking the class into account, their debiasing results are not affected by the number of classes and perform best for this data set.

\section{Regression}
\label{sec:regression_task}

We finally turn to the regression setting. The task is to predict the valence (sentiment) of a given Facebook post, where each post is assigned a valence score by two trained annotators in the range 1--9 and the task is to predict the average of the two scores \cite{Preoctiuc:16}. Additionally, each post is associated with a binary authorship gender label.\footnote{This dataset is also annotated with arousal scores but corresponding results are less biased, and as a result, we focus on bias  mitigation for valence predictions. Results for arousal predictions are included in \Cref{valence_appendix}.

} In our experiments, results are reported based on 5-fold cross-validation. 
 
\subsection{Results}

Instead of measuring fairness with \gap based on TPR scores for classification tasks, we focus on the Pearson correlation disparities across demographic groups.
From \Cref{valence_results} we can see that all models
improve over \vanilla.
Overall, \inlp is the best debiasing method, which we hypothesise is because its linear structure is more appropriate for the small data set, while the deeper methods appear to overfit.

\begin{table}
\centering
\small
	\begin{tabular}{lrrr}
		\toprule
		Models &  \pearson $\uparrow$  &  \fairness$\uparrow$ & \dto$\downarrow$ \\
		\midrule
            \vanilla &        63.38$\pm$2.48 &     85.18$\pm$0.40 &  39.50 \\
             \rw     &        63.69$\pm$1.50 &     84.73$\pm$0.91 &  39.39 \\
                \inlp &        \bf 70.46$\pm$0.00 &     88.54$\pm$0.00 &  \bf 31.68 \\
                 \adv &        69.41$\pm$0.39 &     85.81$\pm$0.33 &  33.72 \\
                \dadv &        69.02$\pm$0.85 &     85.66$\pm$0.63 &  34.14 \\
            \fairbatch &        68.25$\pm$1.47 &     85.18$\pm$0.62 &  35.04 \\
            \difference &        65.88$\pm$0.89 &     85.05$\pm$0.40 &  37.25 \\
            \mean    &        65.37$\pm$1.29 &     85.03$\pm$0.39 &  37.73 \\

		\bottomrule
	\end{tabular}
\caption{Experimental results on the Valence test set. }
\label{valence_results}
\end{table}


\section{\cameraReady{General Discussion and Recommendations}}
\label{sec:resampling_and_reweighting_methods}

So far, we have shown that there is no single best model across different data conditions, and data conditions should be a key consideration in fairness evaluation. 
In this section, we divide debiasing methods into three families, and summarize their robustness to skewed training data distributions.

\paragraph{Balancing demographics in the training dataset}
\ds and \rw are representatives of this family, and are simple and effective. 
In addition, such methods are flexible as the training dataset is pre-processed before model training, and any candidate models on the original dataset can be applied to the debiased dataset.

However, \ds methods are sensitive to group sizes. Considering an extreme setting where the smallest subset in the training dataset has 0 instances, i.e., $\train_{\ervc,\ervg}=\emptyset$, \ds will result in an empty training set. For instance, the group size distribution is highly skewed for the regression task, and \ds resulted in $r=0$ Pearson correlation (\Cref{tab:valanece} in Appendix).
Similar problems are associated with up-sampling methods, which can increase the training set size dramatically.

In addition, when considering multiple protected attributes, such as intersectional groups and gerrymandering groups~\citep{Subramanian:21b}, the number of groups increases exponentially with the number of protected attributes to be considered. 
As a result, the joint distributions can be highly skewed, and these two families of methods (resampling and reweighting) may not be appropriate choices.

Lastly, skewed protected label distributions in the training dataset is not the only source of bias~\citep{Wang:19}. 
For example, as shown in \Cref{class_ratio_5_gender} the \vanilla model trained over balanced versions ($P'(\aae|\happy) = 0.5$) of the \moji dataset is less fair than the \vanilla model trained over a biased dataset where $P'(\aae|\happy) = 0.4$.

\paragraph{Learning fair hidden representations}
\adv, \dadv, and \inlp represent a family of methods that learn fair representations through unlearning discriminators. 
Since the training and unlearning of discriminators do not take into account target class information, these methods are robust to the number of classes and naturally generalize to regression tasks.

However, these methods are not capable of modelling conditional independence for the equal opportunity criterion without taking target class into consideration, resulting in worse \dto than other debiasing methods over \moji (\Cref{sec:binary_classification}).
To achieve equal opportunity fairness, different discriminators can be trained for each target class to capture conditional independence~\citep{Shauli:20,Han2022TowardsEO}. 
But training target-specific discriminators assumes target labels to be discrete, which is sensitive to the number of classes.

Another limitation of this family of methods is associated with the discriminator learning: the discriminator can also suffer from long-tail learning problems, i.e.\ skewed demographics, and lead to biased estimations of protected information.
The unlearning of biased discriminators limits the method's contribution to bias mitigation, which can be seen from Figures~\ref{class_ratio_9_gender} and~\ref{gender_ratio_9_class} in \Cref{sec:binary_classification}.


\paragraph{Minimising loss disparities across demographic groups}
\fairbatch, \difference, and \mean provide a practical approximation of expected fairness in using empirical risk-based objectives, and directly optimize for empirical risk parity during training.

Similar to balanced training approaches, resampling and reweighting are also used in mitigating loss disparities, where \fairbatch adjusts resampling probabilities for batch selection, and \difference and \mean assign instances different weights depending on the demographic group they belong to.
However, minimising loss disparities can be more flexible than balanced training -- for example, instance weights are dynamically adjusted by \difference and \mean, and can take on negative values to aggressively reduce a bias towards favouring of over-represented groups.

Conversely, drawbacks associated with resampling and reweighting also apply to this family.
For example, \fairbatch indeed broke down (an error raised) when $\train_{\ervc,\ervg}=\emptyset$ for the minority group in a particular minibatch for a \bios dataset variant where the smallest group size is close to 0 (\Cref{sec:profession_classification}).

Minimising loss difference is also less efficient in multi-class settings, as it adjusts weights based on class information during training, making optimisation harder.

\section{Conclusion}

In this work, we presented a novel framework for investigating different classification dataset distributions with a single parameter, and used it to systematically examine the effectiveness of debiasing methods in binary classification and multi-classification settings based on real-world datasets. We also presented preliminary analysis of debiasing methods in a regression setting, including proposing a method for adapting existing debiasing methods to regression tasks.
Based on extensive experimentation over three datasets, we found that there was no single best model. Debiasing methods that account for both class and demographic disparities are generally more robust, but are less efficient at achieving fairness in multi-class settings. 
For the regression task, we demonstrated that existing debiasing approaches can substantially improve fairness, and that the simple linear debiasing method outperforms more complex techniques.
In summary, there is no universal best debiasing method across all tasks, and data conditions have a large impact on different models. As such, we propose that future research adopts our evaluation framework as a means of more comprehensively evaluating debiasing methods.

\section*{Acknowledgments}
We thank the anonymous reviewers for their helpful feedback and suggestions. 
This work was funded by the Australian Research Council, Discovery grant DP200102519.
This research was undertaken using the LIEF HPC-GPGPU Facility hosted at the University of Melbourne. This Facility was established with the assistance of LIEF Grant LE170100200.

\section*{\cameraReady{Limitations}}
This paper focuses on fairness evaluation \wrt equal opportunity fairness. While a more comprehensive study should include a diversity of fairness objectives, we note that previous work~\citep{Han:21} has shown that evaluation results \wrt different fairness criteria are highly correlated. 

Consistent with previous work, we restrict our experiments to  categorical protected attributes (binary gender, ethnicity) acknowledging that other relevant attributes (such as age) are more naturally modeled as a continuous variable. Since the aim of this paper is a systematic evaluation of existing debiasing methods, which were all developed specifically for categorical protected attributes, the extension to continuous variables is beyond the scope of this paper. 
A simple adaptation to continuous demographic labels like age is discretization, which we leave as a promising direction for future work.

For similar reasons, we use established data sets as provided by the original authors and used in relevant prior work, and acknowledge the simplified treatment of gender as a binary variable which reflects neither the diversity nor the fluidity of the underlying concept~\cite{dev-etal-2021-harms}.

\section*{Ethical Consideration}

In this work, we focus on examining the effectiveness of various debiasing methods on both classification and regression tasks, where the protected attribute is either ethnicity or gender. However, their effectiveness in reducing bias towards other protected attributes is not necessarily guaranteed. Furthermore, the protected attributes examined in our work are limited to binary labels, whose effectiveness in debiasing {$N$}-ary protected attributes are left to future work.

\bibliographystyle{acl_natbib}
\bibliography{acl2022}

\clearpage
\appendix

\section{Datasets and Implementation Details}
\label{sec:implementation_details}

\subsection{\moji}
Following previous studies \cite{Shauli:20,Han:21}, the original training dataset is balanced with respect to both sentiment and ethnicity but skewed in terms of sentiment--ethnicity combinations (40\% \happy-\aae, 10\% \happy-\sae, 10\% \sad-\aae, and 40\% \sad-\sae, respectively). Note that the dev and test set are balanced in terms of sentiment--ethnicity combinations.
The dataset contains 100K/8K/8K train/dev/test instances.

When varying training set distributions, we keep the 8k test instances unchanged.

We use \deepmoji \cite{Felbo:17} to obtain Twitter representations, where \deepmoji is a model pretrained over 1.2 billion English tweets and \deepmoji is fixed during model training. For all models, the learning rate is 3e-3, and the batch size is 1,024. Hyperparameter tuning for each model is described in \Cref{mojj_original_hyperparameters}. 

\begin{table}[!t]
    \centering
    \small
    \begin{tabular}{lrrrr}
\toprule
       Profession &  Total &  Male &  Female &  Ratio \\
\midrule
        dietitian &              2567 &              183 &               2384 &         0.929 \\
            nurse &             12316 &             1127 &              11189 &         0.908 \\
        paralegal &              1146 &              173 &                973 &         0.849 \\
     yoga\_teacher &              1076 &              166 &                910 &         0.846 \\
            model &              4867 &              840 &               4027 &         0.827 \\
interior\_designer &               949 &              182 &                767 &         0.808 \\
     psychologist &             11945 &             4530 &               7415 &         0.621 \\
          teacher &             10531 &             4188 &               6343 &         0.602 \\
       journalist &             12960 &             6545 &               6415 &         0.495 \\
        physician &             26648 &            13492 &              13156 &         0.494 \\
             poet &              4558 &             2323 &               2235 &         0.490 \\
          painter &              5025 &             2727 &               2298 &         0.457 \\
 personal\_trainer &               928 &              505 &                423 &         0.456 \\
        professor &             76748 &            42130 &              34618 &         0.451 \\
         attorney &             21169 &            13064 &               8105 &         0.383 \\
       accountant &              3660 &             2317 &               1343 &         0.367 \\
     photographer &             15773 &            10141 &               5632 &         0.357 \\
          dentist &              9479 &             6133 &               3346 &         0.353 \\
        filmmaker &              4545 &             3048 &               1497 &         0.329 \\
     chiropractor &              1725 &             1271 &                454 &         0.263 \\
           pastor &              1638 &             1245 &                393 &         0.240 \\
        architect &              6568 &             5014 &               1554 &         0.237 \\
         comedian &              1824 &             1439 &                385 &         0.211 \\
         composer &              3637 &             3042 &                595 &         0.164 \\
software\_engineer &              4492 &             3783 &                709 &         0.158 \\
          surgeon &              8829 &             7521 &               1308 &         0.148 \\
               dj &               964 &              828 &                136 &         0.141 \\
           rapper &               911 &              823 &                 88 &         0.097 \\
\bottomrule
\end{tabular}
    \caption{Statistics of the \bios training dataset. Ratio stands for the percentage of female individuals for each profession}
    \label{tab:bios_distribution}
\end{table}

\subsection{\bios}
\label{sec:appendix_bios_distribution}

We denote the data set as \bios, and use the same split as prior work~\cite{Shauli:20,Shen:22} of 257k train, 40k dev and 99k test instances.
\Cref{tab:bios_distribution} shows the number of instances of each profession, the number of male and female individuals of each profession, and the ratio of female individuals for each profession in the \bios training dataset.
As the target label distribution is highly skewed, we adjust the distribution over \bios dataset with 30K training instances, such that each profession contains about 1K instances, which is similar to the size of the smallest target group.

We use the ``CLS'' token representation of the pretrained uncased \bert-base \cite{Devlin:19} to obtain text representations, where \bert-base is fixed during model training, aligning with previous studies \cite{Shauli:20,Shen:22}. Hyperparameter settings for all models are available in \Cref{bios_original_hyperparameters}.

\subsection{Valence}
The dataset contains 2,883 posts, of which male and female authors account for $51\%$ and $49\%$ respectively. 

We use the ``CLS'' token representation of the pretrained uncased \bert-base \cite{Devlin:19} to obtain post representations, where \bert-base is fixed during model training. Hyperparameter settings are described in \Cref{valence_hyperparameters}.

For this task, we use \pearson, mean absolute error (\mae), and root mean square error (\rmse) to evaluate model performance; and we use the Pearson difference (\pearsongap), \mae difference (\maegap), and \rmse difference (\rmsegap) between male and female groups to evaluate model bias.

\section{Normalization For Probability Table}
\label{sec:appendix_Normalization_For_Probability_Table}

To make sure the resulting probability table $P'$ is valid, we normalize the table by replacing negative values with 0, and normalize the sum to 1.
Specifically, let $S = \sum_\ervy\sum_\ervz P'(\ervy, \ervz)$ denote the sum of probabilities. The normalization is $P'(\ervy, \ervz) = \frac{P'(\ervy, \ervz)}{S}, \forall \ervy, \ervz$.

\section{Twitter Sentiment Analysis}
\label{moji_appendix}

\subsection{Hyperparameters}
\label{mojj_original_hyperparameters}

For all models except for \vanilla, \ds, and \rw, where no extra hyperparameters are introduced, we tune the most sensitive hyperparameters through grid search. For \inlp, following \citet{Shauli:20}, we use $300$ linear SVM classifiers. For \adv, we tune $\lambda_{\text{adv}}$ from 1e-3 to 1e3 with 60 trials. For \dadv, we further tune $\lambda_{\text{diverse}}$ within the range of 1e-1 and 1e5 with 60 trials. For \fairbatch, we tune $\alpha$ from 1e-3 to 1e1 with 40 trials. For \difference and \mean, we tune $\lambda$ within the range of 1e-3 and 1e1 with 40 trials, respectively. All hyperparameters are finetuned on the \moji dev set.

\subsection{Results}
\label{moji_result}

\begin{table}
	\scalebox{0.93}{
		\begin{tabular}{lccc}
			\toprule
			Models &  \accuracy$\uparrow$  &  \fairness$\uparrow$    &\dto$\downarrow$ 
			\\
			\midrule
			\vanilla &                  72.49$\pm$0.18 &               60.79$\pm$1.12 &  47.90  
			\\
			\ds &                  \textbf{75.92}$\pm$0.32 &               86.88$\pm$1.08 &  27.43 
			\\
			
			\btrw &                  \textbf{75.96}$\pm$0.28 &               86.18$\pm$0.97 &  27.73  
			\\
			\inlp &                  73.18$\pm$0.00 &               82.04$\pm$0.00 &  32.28  
			\\
			\adv &                  75.12$\pm$0.83 &               90.40$\pm$1.75 &  26.67 
			\\
			\dadv &                  75.65$\pm$0.12 &               89.94$\pm$0.50 &  \textbf{26.34}  
			\\
			\fairbatch &                  74.96$\pm$0.41 &               90.49$\pm$0.49 &  26.79  
			\\
			\difference &                  75.09$\pm$0.25 &               \textbf{90.70}$\pm$0.87 &  26.59  
			\\
			\mean &                  75.60$\pm$0.17 &               89.83$\pm$0.60 &  26.43  
			\\
			\bottomrule
		\end{tabular}
	}
	\caption{Experimental results on the \moji test set (averaged over 5 runs); \textbf{Bold} $=$ Best Performance; $\uparrow =$ the higher the better; $\downarrow =$ the lower the better. }
	\label{moji_table_result}			
\end{table}

\Cref{moji_table_result} shows the results achieved by various methods. All debiasing methods can reduce bias significantly while improving model performance in terms of \accuracy. 

\section{Profession Classification}
\label{bios_appendix}

\subsection{Hyperparameters}
\label{bios_original_hyperparameters}

For all models, the learning rate is 3e-3, and the batch size is 1,024. For all models we tune the most sensitive hyperparameters through grid search except for \vanilla, \ds, and \btrw as there is no extra hyperparameters introduced for these three methods. For \inlp, following \citet{Shauli:20}, we use $300$ linear SVM classifiers. For \adv, we tune $\lambda_{\text{adv}}$ from 1e-3 to 1e3 with 60 trials. For \dadv, we further tune $\lambda_{\text{diverse}}$ within the range of 1e-1 and 1e5 with 60 trials. For \fairbatch, we tune $\alpha$ from 1e-3 to 1e1 with 40 trials. For \difference and \mean, we tune $\lambda$ within the range of 1e-3 and 1e1 with 40 trials, respectively. All hyperparameters are finetuned on the \bios dev set.

\section{Adaptation For Regression Tasks}
\label{sec:appendix_regression_adaptation}

\subsection{\difference~\citep{Shen:22}}
The debiasing objective for classification tasks is to minimise cross-entropy loss disparities across different protected groups within each class, 
$\mathcal{L}_{\text{eo}}^{\text{class}} = \lambda\sum_{\ervc=1}^{\ermC}\sum_{\ervg =1}^{\ermG}|\mathcal{L}^{\ervc,\ervg}_{ce}-\mathcal{L}_{ce}^{\ervc}|$,
where $\mathcal{L}^{\ervc,\ervg}_{ce}$ and $\mathcal{L}_{ce}^{\ervy}$ are the cross-entropy losses for subset of instances $\{(\vx_{i}, \ervy_{i}, \ervz_{i})|\ervy_{i}=\ervc, \ervz_{i}=\ervg\}_{i=1}^{n}$ and $\{(\vx_{i}, \ervy_{i}, \ervz_{i})| \ervy_{i}=\ervc\}_{i=1}^{n}$, respectively.

Clearly, the identification of subsets requires categorical labels, which is based on proxy labels for regression tasks. By replace the cross-entropy loss with mean squared error loss ($\mathcal{L}_{mse}$), the objective for \difference is 
$\mathcal{L}_{\text{eo}}^{\text{reg}} = \lambda\sum_{\tilde{\ervc}=1}^{\tilde{\ermC}}\sum_{\ervg =1}^{\ermG}|\mathcal{L}^{\tilde{\ervc},\ervg}_{mse}-\mathcal{L}_{mse}^{\tilde{\ervc}}|$
where $\mathcal{L}^{\tilde{\ervc},\ervg}_{mse}$ and $\mathcal{L}_{mse}^{\tilde{\ervc}}$ are the cross-entropy losses for subset of instances $\{(\vx_{i}, \ervy_{i}, \ervz_{i}, \tilde{\ervy}_{i})|\tilde{\ervy}_{i}=\tilde{\ervc}, \ervz_{i}=\ervg\}_{i=1}^{n}$ and $\{(\vx_{i}, \ervy_{i}, \ervz_{i}, \tilde{\ervy}_{i})| \tilde{\ervy}_{i}=\tilde{\ervc}\}_{i=1}^{n}$, respectively.

\begin{table*}[!t]
	\scalebox{0.97}{ 
		\begin{tabular}{lcccccc}
			\toprule
			Models &  \pearson$\uparrow$ & \pearsongap$\downarrow$ &  \mae$\downarrow$ & \maegap$\downarrow$ &\rmse$\downarrow$  & \rmsegap$\downarrow$ 
			\\
			\midrule
			\vanilla &0.63$\pm$0.04  &\textbf{0.06}$\pm$0.05      &0.78$\pm$0.03      &0.08$\pm$0.01  &1.00$\pm$0.04  &0.09$\pm$0.02
			\\
			\ds  &0.00$\pm$0.04  &0.08$\pm$0.04  &0.97$\pm$0.05  &0.06$\pm$0.03  &1.23$\pm$0.05  &0.05$\pm$0.03
			\\
			\btrw & 0.62$\pm$0.03  &\textbf{0.06}$\pm$0.05  &0.78$\pm$0.02  &0.08$\pm$0.02  &0.99$\pm$0.03  &0.09$\pm$0.04
			\\
			\inlp  & 0.66$\pm$0.04  & 0.09$\pm$0.04  &\textbf{0.71}$\pm$0.04  &\textbf{0.03}$\pm$0.02  &\textbf{0.92}$\pm$0.04  &\textbf{0.04}$\pm$0.02
			\\
			\adv  & \textbf{0.67}$\pm$0.03  &\textbf{0.06}$\pm$0.06  &0.72$\pm$0.03  &0.06$\pm$0.04  &0.93$\pm$0.04  &0.09$\pm$0.06
			\\
			\dadv  &\textbf{0.67}$\pm$0.03  & 0.07$\pm$0.06  &0.72$\pm$0.02  &0.06$\pm$0.02  &\textbf{0.92}$\pm$0.02  &0.07$\pm$0.05
			\\
			\fairbatch & \textbf{0.67}$\pm$0.03 &\textbf{0.06}$\pm$0.06  &\textbf{0.71}$\pm$0.01   &0.06$\pm$0.02  &\textbf{0.92}$\pm$0.02  &0.07$\pm$0.04
			\\
			\difference  &0.65$\pm$0.03  & 0.07$\pm$0.05  &0.75$\pm$0.03  &0.07$\pm$0.01  &0.96$\pm$0.03  &0.08$\pm$0.02
			\\
			\mean &0.64$\pm$0.03  &\textbf{0.06}$\pm$0.06  &0.76$\pm$0.03 &0.08$\pm$0.02  &0.97$\pm$0.04  & 0.10$\pm$0.04
			\\
			\bottomrule
	\end{tabular}}
	\caption{Experimental results on the Facebook post dataset with respect to arousal; the best performance is indicated in bold.}
	\label{tab:valanece}
\end{table*}

\section{Arousal Prediction of Facebook Posts}
\label{valence_appendix}

\subsection{Hyperparameters}
\label{valence_hyperparameters}

For all models, the learning rate is 7e-4, the batch size is 64, the number of hidden layers is 1, and hidden layer size is 200. Each model is trained with mean squared loss with a weight decay of 1e-3. For all models except for \vanilla, we need to bin instances,  as the dataset is small and the range of valence scores is large; otherwise, these methods cannot be applied in their original form. In this work, instances are grouped into 4 bins. For all models we tune the most sensitive hyperparameters through grid search except for \vanilla, \ds, and \btrw as there are no extra hyperparameters introduced for these three methods. For \inlp, following \citet{Shauli:20}, we use $200$ linear regressors. For \adv, we tune $\lambda_{\text{adv}}$ from 1e-3 to 1e3 with 60 trials. For \dadv, we further tune $\lambda_{\text{diverse}}$ within the range of 1e-1 to 1e5 with 60 trials. For \fairbatch, we tune $\alpha$ from 1e-3 to 1e1 with 40 trials. For \difference and \mean, we tune $\lambda$ within the range of 1e-3 to 1e1 with 40 trials, respectively. All hyperparameters are finetuned on the dev set. 

\subsection{Results}
\Cref{tab:valanece} presents the results on the arousal dataset.

\end{document}